\documentclass{article}
\usepackage{spconf,amsmath,graphicx,caption}
\usepackage{amssymb}
\usepackage{multirow}
\usepackage{subfig}
\usepackage{wrapfig}
\usepackage{hyperref}
\usepackage{cleveref}
\usepackage{booktabs}
\usepackage{float}
\usepackage{enumitem}
\usepackage[dvipsnames,table]{xcolor}

\title{Freeze the backbones: A Parameter-Efficient Contrastive Approach to Robust Medical Vision-Language Pre-training}
\name{Jiuming Qin $^{1*}$, \qquad Che Liu$^{2,3*}$, \qquad Sibo Cheng$^{1,2}$, \qquad Yike Guo$^{2,4}$, \qquad Rossella Arcucci$^{2,3}$ \thanks{$^*$ Equal contribution.}}
\address{$^{1}$ Department of Computing, Imperial College London\\
$^{2}$ Data Science Institute, Imperial College London\\
$^{3}$ Department of Earth Science and Engineering, Imperial College London\\
$^{4}$ Department of Computer Science and Engineering, The Hong Kong University of Science and Technology
}
\begin{document}
% \ninept
%
\maketitle
\begin{abstract}
Modern healthcare often utilises radiographic images alongside textual reports for diagnostics, encouraging the use of Vision-Language Self-Supervised Learning (VL-SSL) with large pre-trained models to learn versatile medical vision representations. However, most existing VL-SSL frameworks are trained end-to-end, which is computation-heavy and can lose vital prior information embedded in pre-trained encoders. To address both issues, we introduce the backbone-agnostic Adaptor framework, which preserves medical knowledge in pre-trained image and text encoders by keeping them frozen, and employs a lightweight Adaptor module for cross-modal learning. Experiments on medical image classification and segmentation tasks across three datasets reveal that our framework delivers competitive performance while cutting trainable parameters by over $90\%$ compared to current pre-training approaches. Notably, when fine-tuned with just $1\%$ of data, Adaptor outperforms several Transformer-based methods trained on full datasets in medical image segmentation.   
\end{abstract}
\section{Introduction}
Recent advances in deep learning that combine visual and textual data have achieved significant breakthroughs in the medical field, where available data frequently pairs visual elements, such as radiographs and CT scans, with textual clinical reports \cite{tsimpoukelli2021multimodal, radford2021learning}. As a result, Vision-Language Self-Supervised Learning (VL-SSL) has emerged as a prevalent pre-training strategy to learn medical representations. It leverages the expansive medical dataset of paired multi-modal data to reduce the dependency on data annotations that traditional supervised training demands \cite{oord2019representation, chen2020simple}. 
Combined with the rapid developments of domain-specific pre-trained models, researchers can now harness data from multiple modalities to enhance performance in both uni-modal and multi-modal tasks.
Existing works in this field primarily focus on fusing cross-modal information by \textit{aligning} vision and text features with end-to-end training \cite{su2020vlbert,wang2022multi,M_ller_2022,zhang2022contrastive}. However, two main challenges arise: (i) the high computational cost and unstable training associated with jointly modelling two distinct modalities with large-sized backbone models \cite{zhai2022scaling,chen2021empirical,wang2022one}, and (ii) the risk of diluting prior in-domain information embedded in pre-trained backbones due to finetuning, under-utilising their respective power \cite{Boecking_2022,qin2023medical}. 
Previous studies have assessed the benefits of partially freezing components within vision-language pre-training frameworks, focusing on either language models \cite{liu2023mflag} or vision backbones \cite{qin2023medical}, but not both simultaneously. Recognising that individually freezing sections of language models or vision backbones helps medical image understanding, we investigate the effects of simultaneously freezing both components completely, and present a robust and efficient vision-language pre-training method based on this approach. Our key contributions are as follows:
\begin{itemize}[leftmargin=*,topsep=1pt]
    \itemsep0em
    \item We introduce the Adaptor framework with a lightweight, trainable module that effectively integrates knowledge from frozen medical image and text encoders through self-supervised learning. This approach offers notable advantages in parameter efficiency and low computational cost compared to traditional end-to-end pre-training. 
    \item On medical image classification and segmentation tasks over three datasets, Adaptor delivers competitive results while utilising over 90\% fewer trainable parameters than current methods. Notably, when fine-tuned on 1\% of the data, Adaptor surpasses multiple Transformer-based techniques trained on complete datasets for medical image segmentation.
\end{itemize}

\section{Methodology}
\begin{figure*}[htb!]
  \centering
  \includegraphics[width=.95\textwidth]{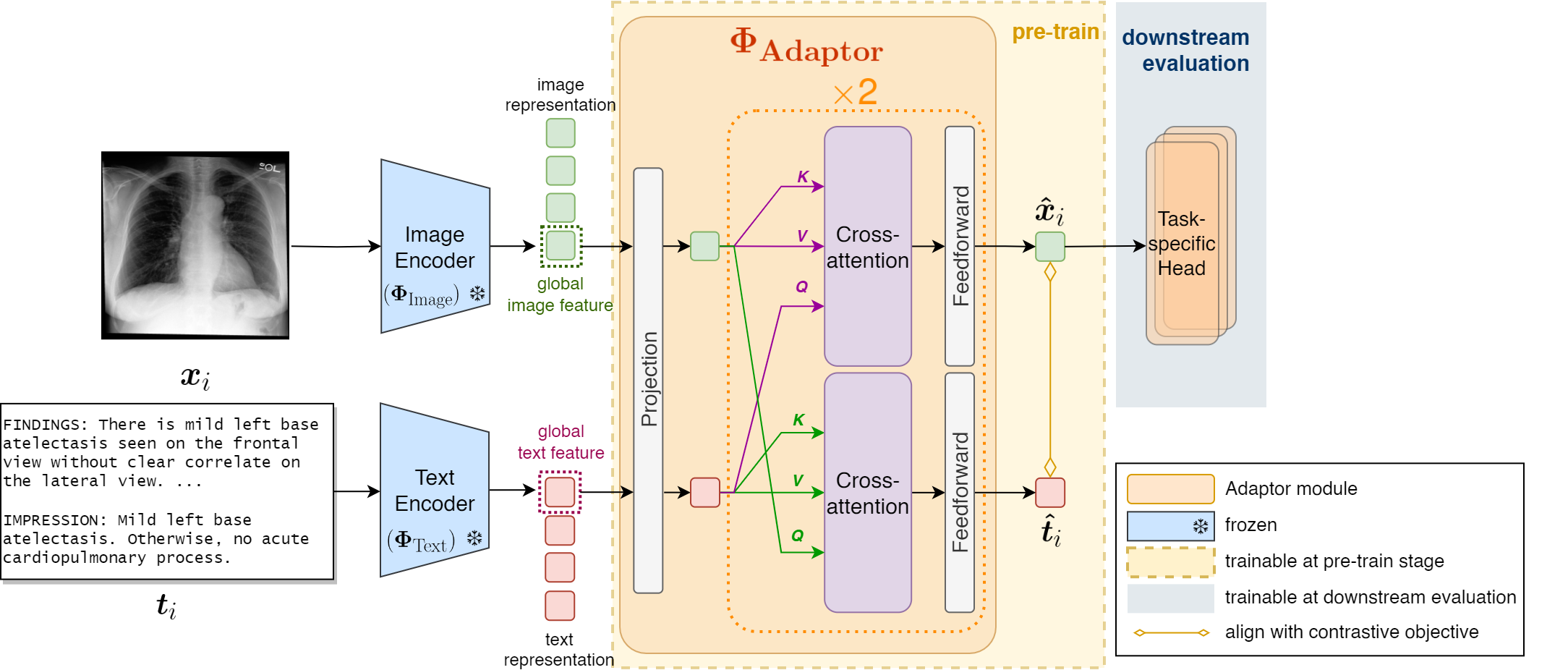}
  \caption{{The Adaptor framework. {Note that the duplicated cross-attention and feedforward blocks are identical, only shown this way to demonstrate the different choices of KVQ vectors in the attention mechanism for two modalities. \textit{Blue model blocks} are frozen during both pre-train and finetune, while \textit{yellow and grey blocks} are updated during pre-train and downstream task evaluation stages respectively. }}}
  \label{fig:architecture}
\end{figure*}

\subsection{Problem Definition and Framework}
Given a training dataset $\mathcal D_{train} = \big\{(\mathcal {\boldsymbol{x}}_i, \mathcal {\boldsymbol{t}}_i)\big\}_{i=1}^N$, consisting of  corresponding radiographic images and textual reports, our goal is to train a model that jointly processes and understands visual and textual information, resulting in robust medical visual representations. We utilise  pre-trained, frozen vision and language backbones to extract uni-modal representations, and introduce an Adaptor module to facilitate cross-modal learning. 
\Cref{fig:architecture} illustrates our Adaptor framework, which comprises of:
\begin{itemize}[leftmargin=*]
    \itemsep0em
\item \textbf{Frozen Pre-trained Vision and Language Backbones $(\Phi_\text{image}, \Phi_\text{text})$:} These transform raw images and texts into individual uni-modal embeddings 
$\big\{(\Phi_\text{image}(\mathcal {\boldsymbol{x}}_i), \Phi_\text{text}(\mathcal {\boldsymbol{t}}_i))\big\}_{i=1}^N$. 
\item \textbf{Adaptor module $\Phi_{\text{Adaptor}}$:} This module projects the vision and text embeddings onto the same dimensionality via a linear layer, then employs two Transformer layers with {cross-attention mechanisms} to generate multi-modal-aware embeddings, \(\hat{\boldsymbol{x}}_i\) and \(\hat{\boldsymbol{t}}_i\).
 The resulting image embeddings $\{\hat{\mathcal {\boldsymbol{x}}}_i\}$ are then utilised directly for downstream tasks.  
\end{itemize}

%%%%%%%%%%%%%%%%%%%%%%%%%%%%%%%%%%%%%%%%%%%%%%%%%%%%%%%%%%%%%%%%%%%%%%%%%%%%%%%%%%%%%%

\subsection{Pre-training Objective}
The pre-training objective jointly models the embeddings of image-text pairs in a shared latent space by minimising a contrastive loss, which enhances the correlation between corresponding image-text pairs and separate the unrelated ones.  
We sample batches of $n$ image-text pairs, pass them though the frozen dual encoders and the trainable Adaptor module, to obtain multi-modal-aware image embeddings $\hat{\mathcal{X}} = \{\hat{\boldsymbol{x}}_1, ..., \hat{\boldsymbol{x}}_n\}
$ and text embeddings $\hat{\mathcal{T}} = \{\hat{\boldsymbol{t}}_1, ..., \hat{\boldsymbol{t}}_n\}$:
\begin{equation}
    (\hat{\mathcal {\boldsymbol{x}}}_i, \hat{\mathcal {\boldsymbol{t}}}_j) = \Phi_{\text{Adaptor}}\left[(\Phi_{\text{image}}(\mathcal {\boldsymbol{x}}_i), \Phi_{\text{text}}(\mathcal {\boldsymbol{t}}_j)\right], \quad i,j=1, ..., n.
\end{equation}
The image-to-text contrastive loss \(\mathcal L_{\textsc{I-T}}\), which takes a similar form as the InfoNCE loss \cite{oord2019representation}, is given by:
\begin{equation}
\mathcal L_{\textsc{I-T}} = -\frac{1}{n} \sum_{k=1}^{n} \log \frac{\exp\{\langle\hat{x}_k, \hat{t}_k\rangle/\tau\}}{\sum_{j=1}^{N} \exp\{\langle \hat{x}_k, \hat{t}_j\rangle/\tau\}},
\end{equation}
where $\langle \cdot, \cdot \rangle$ denotes inner product. This loss encourages the model to increase the similarity between the matching image-text pairs \( \{(\hat x_k, \hat t_k)\}_{k=1}^n \) and reduce similarities associated with the other text samples in the minibatch. $\tau$ is a learnable temperature parameter to further differentiate between positive and negative samples, shown to be effective in enabling the models to learn from hard negatives and proved necessary in vision self-supervised learning \cite{chen2020simple}.  

By symmetry, the text-to-image loss, \(\mathcal L_{\textsc{T-I}}\), is given by:
\begin{equation}
\mathcal L_{\textsc{T-I}} = -\frac{1}{n} \sum_{k=1}^{n} \log \frac{\exp\{\langle\hat{x}_k, \hat{t}_k\rangle/\tau\}}{\sum_{j=1}^{N} \exp\{\langle \hat{x}_j, \hat{t}_k \rangle/\tau\}}.
\end{equation}
The final loss $\mathcal L$ is a weighted sum of the two:
\begin{equation}
\label{eq:final-loss}
\mathcal L = \alpha\mathcal L_{\textsc{I-T}} + (1-\alpha)\mathcal L_{\textsc{T-I}},
\end{equation}
where $\alpha\in(0, 1)$ is a pre-determined weight hyperparameter.

Unlike previous vision-language learning methods that rely the deterministic weight $\alpha$ to introduce asymmetry between textual and visual loss terms \cite{zhang2022contrastive,M_ller_2022},
our architecture and loss design provide additional flexibility in attending to information from each modality. Specifically, the cross-attention mechanism in the Adaptor module dynamically adjusts the influence of each modality based on training data, offering an implicit, context-aware weighting scheme. 

% \section{Experiments and Analysis}
\section{Experiments}
To demonstrate the flexibility and robustness of the Adaptor framework, we evaluated its performance using different combinations among three pre-trained image encoders, namely ResNet autoencoder \cite{cohen2020limits,Cohen2022xrv}, DINOv2-small and DINOv2-base \cite{oquab2023dinov2}, and five pre-trained language models, which are BERT \cite{devlin2019bert}, BioBERT \cite{lee2020biobert}, ClinicalBERT \cite{huang2020clinicalbert}, CXR-BERT \cite{Boecking_2022} and PubMedBERT \cite{pubmedbert}. 

% \subsection{Dataset and Experimental Setup}
\noindent\textbf{Pre-train.}
We pre-train the Adaptor framework with the MIMIC Chest X-ray Database v2.0.0 \cite{mimic-cxr-2}, which consists of $277,110$ chest radiographs paired with $227,835$ medical reports. We adopt the original dataset splits, using the training data for model updates and both validation and test data for model validation. We set the weight hyperparameter $\alpha$ in \Cref{eq:final-loss} to be 0.75 as per Zhang et al. \cite{zhang2022contrastive}, and train the model for 50 epochs with a batch size of 1024 and a learning rate of 2e-5. 
Since most parameters in the framework remain frozen, we pre-compute and store all output embeddings from frozen encoders in advance, and load them directly as inputs during Adaptor pre-training. This significantly accelerates training, which takes approximately 15 minutes to run on two Tesla T4 GPUs. 

\noindent\textbf{Downstream Tasks.}
We evaluated the framework on two tasks: medical image classification and medical image segmentation. In this stage, both the dual encoders and the pre-trained Adaptor module are kept frozen, as shown in \Cref{fig:architecture}. Experiments were conducted on $1\%$, $10\%$ and $100\%$ of the available labelled data. \textbf{\textit{Medical image classification:}} We use the RSNA Pneumonia \cite{RSNA} and COVIDx CXR-2 datasets \cite{covidx}. The text encoder is detached from the framework due to the lack of textual inputs at this stage, and a two-layer linear classification head is added on top of the Adaptor. We optimise the cross entropy loss, and report the test set AUROC score for RSNA and accuracy for COVIDx dataset. \textbf{\textit{Medical image segmentation:}} The RSNA Pneumonia \cite{RSNA} and SIIM-ACR datasets \cite{siim} are used. We again detach the text encoder and add a decoder after the Adaptor module. For the ResNet Autoencoder, we use a U-Net-style decoder \cite{ronneberger2015unet}, while for the DINOv2 Vision Transformer, an upsampling convolutional decoder is chosen due to the lack of layers in the Transformer architecture. 

\section{Analysis}
% \subsection{Analysis}
\textbf{Backbone Compatibility.} The Adaptor module is widely compatible with different image encoder architectures. In addition to Vision Transformers which naturally yield a global feature, convolutional encoders like ResNet are also compatible when using the channel-wise average of the output feature map as the global uni-modal representation. In \Cref{tab:classification-rsna-auroc}, we show the performance of Adaptors using different backbone models, each fine-tuned with sections of the RSNA dataset. The consistent performance across vision models underscores the robustness of the Adaptor module when integrated with diverse backbone architectures.
\begin{table}[htb]
\footnotesize
\centering
\caption{Classification AUROC score [\%] of Adaptor with different vision and language models on RSNA dataset.}
\vspace*{-3mm}
\label{tab:classification-rsna-auroc}
\setlength{\tabcolsep}{3pt}
\begin{tabular}{@{}l|ccc|ccc|ccc@{}}
\toprule
\textbf{}                             & \multicolumn{3}{c|}{ResNet-AE} & \multicolumn{3}{c|}{DINOv2-S} & \multicolumn{3}{c}{DINOv2-B}                  \\
\multicolumn{1}{r|}{\textit{data \%}} & 1\%      & 10\%     & 100\%    & 1\%      & 10\%    & 100\%    & 1\%           & 10\%          & 100\%         \\ \midrule
BERT                                  & 77.0     & 79.0     & 81.3     & 81.3     & 84.8    & 86.1     & 82.8          & 85.3          & {86.8} \\
BioBERT                               & 76.8     & 79.1     & 81.2     & 82.2     & 85.1    & 86.1     & {84.4} & {85.6} & 86.7          \\
ClinicalBERT                          & 76.7     & 78.9     & 81.1     & 81.7     & 84.9    & 86.1     & 82.9          & 85.3          & 86.6          \\
PubMedBERT                            & 76.3     & 79.0     & 81.1     & 80.9     & 84.5    & 86.0     & 82.3          & 85.0          & 86.5          \\
CXR-BERT                              & 76.9     & 79.3     & 81.5     & 80.8     & 84.7    & 86.0     & 82.3          & 85.1          & 86.6          \\ \bottomrule
\end{tabular}
\end{table}

\begin{table*}[htb!]
\footnotesize
\centering
\caption{Parameter count and performance of existing Vision-Language learning methods on medical image classification.}
\label{tab:compare-clf-params}
\setlength{\tabcolsep}{4pt}
\begin{tabular}{@{}l|lccllllll@{}}
\toprule
 &  & \textbf{Parameters} & \textbf{\textbf{Trainable}} & \multicolumn{3}{c}{RSNA (AUC)} & \multicolumn{3}{c}{COVIDx (ACC)} \\
\textbf{Strategy} & \textbf{Method} & \textbf{(M)} & \textbf{(M)} & 1\% & 10\% & 100\% & 1\% & 10\% & 100\% \\ \midrule
\multirow{2}{*}{\begin{tabular}[c]{@{}l@{}}Baseline\\ (No Transformer)\end{tabular}} & DSVE \cite{engilberge2018finding} & 126.7 & 5.8 & 49.7 & 52.1 & 57.8 & - & - & - \\
 & VSE++ \cite{faghri2018vse} & 11.3 & 11.3 & 49.4 & 57.2 & 67.9 & - & - & - \\ \midrule
\multirow{2}{*}{\begin{tabular}[c]{@{}l@{}}No Medical\\ Pre-training\end{tabular}} & Random Init & 25.6 & 25.6 & 58.9 & 69.4 & 74.1 & 50.5 & 60.3 & 70.0 \\
 & ImageNet Init & 25.6 & 25.6 & 74.9 & 74.5 & 76.3 & 64.8 & 78.8 & 86.3 \\ \midrule
\multirow{5}{*}{\begin{tabular}[c]{@{}l@{}}Medical End-to-End \\ Pre-training\end{tabular}} & GLoRIA-MIMIC \cite{gloria9710099,mimic-cxr-2} & 134.2 & 134.2 & 86.1 & 88.0 & 88.6 & 67.3 & 77.8 & 89.0 \\
 & GLoRIA \cite{gloria9710099} & 134.2 & 134.2 & 86.1 & 88.0 & 88.6 & 67.3 & 77.8 & 89.0 \\
 & ConVIRT \cite{zhang2022contrastive} & 138.1 & 138.1 & 77.4 & 80.1 & 81.3 & 72.5 & 82.5 & 92.0 \\
 & MGCA (ResNet-50) \cite{wang2022multi} & 113.6 & 113.6 & 88.6 & 89.1 & 89.9 & 72.0 & 83.5 & 90.5 \\
 & MGCA (ViT-B) \cite{wang2022multi} & 172.7 & 172.7 & \textbf{89.1} & \textbf{89.9} & \textbf{90.8} & \textbf{74.8} & 84.8 & \textbf{92.3} \\ \midrule
\textbf{\begin{tabular}[c]{@{}l@{}}Medical Frozen-Backbone \\ Pre-training (Ours)\end{tabular}} & \textbf{Adaptor (ViT-B)} & 208.3 & \textbf{{\color{blue}12.2}} & 84.4 & 85.6 & 86.8 & 71.0 & \textbf{87.5} & \textbf{92.3} \\ \bottomrule
\end{tabular}
\end{table*}

\begin{table*}[htb!]
\centering
\footnotesize
\caption{Trainable parameter count and performance on medical image classification. Dice scores [\%] on the test set are reported. }
\vspace*{-5mm}
\setlength{\tabcolsep}{3pt}
% \begin{adjustwidth}{-.5in}{-.5in}  
        \begin{center}
\label{tab:seg-param-perf}
\begin{tabular}{@{}l|cc|rrrrrr@{}}
\toprule
\textbf{\textbf{}}                 & \multicolumn{2}{c|}{\textbf{Trainable Param. (M)}} & \multicolumn{3}{c}{\textbf{RSNA}}                                                & \multicolumn{3}{c}{\textbf{SIIM}}                                                 \\
\textbf{Method}                    & Pre-train              & Decoder              & 1\%                    & 10\%                   & 100\%                  & 1\%                    & 10\%                   & 100\%                  \\ \midrule
{GLoRIA-MIMIC \cite{gloria9710099}}              & 134.2                  & 9.0                  & 60.3	              & 68.7	                & 68.3                      & 37.4                   & 57.1                   & 64.0                   \\
{MGCA (ResNet50-UNet) \cite{wang2022multi}}     & 113.6                  & 9.0                  & 63.0	 & 68.3	 & 69.8 & {49.7} & {59.3} & {64.2} \\
\midrule
\textbf{Adaptor (ResNet-PubMedBERT)}          & \textcolor{red}{12.0}                       & \textcolor{red}{20.1}  & \textcolor{red}{\textbf{76.5}}                       & \textcolor{red}{\textbf{74.3}}  & \textcolor{red}{76.2}  & \textcolor{red}{\textbf{73.1}}                       & \textcolor{red}{\textbf{73.1}}  & \textcolor{red}{73.1}  \\
\textbf{Adaptor (ViT/B-BERT)}                 & {\textcolor{blue}{12.2}} & \textcolor{blue}{10.4} & {\textcolor{blue}{55.9}} & \textcolor{blue}{74.1} & \textcolor{blue}{77.2} & {\textcolor{blue}{51.8}} & \textcolor{blue}{56.8} & \textcolor{blue}{73.3} \\
\textbf{Adaptor (ViT/S-BERT)}                 & {\textcolor{blue}{11.9}} & \textcolor{blue}{10.4} & {\textcolor{blue}{56.0}} & \textcolor{blue}{71.6} & \textcolor{blue}{\textbf{78.1}} &  {\textcolor{blue}{53.1}} & \textcolor{blue}{58.7} & \textcolor{blue}{\textbf{73.7}} \\
 \bottomrule
\end{tabular}
\end{center}
% \end{adjustwidth}
\end{table*}

\begin{figure*}[htb!] 
\centering
\begin{minipage}{.6\textwidth}
  \centering
  \captionsetup{width=.8\linewidth}
    \includegraphics[width=\textwidth]{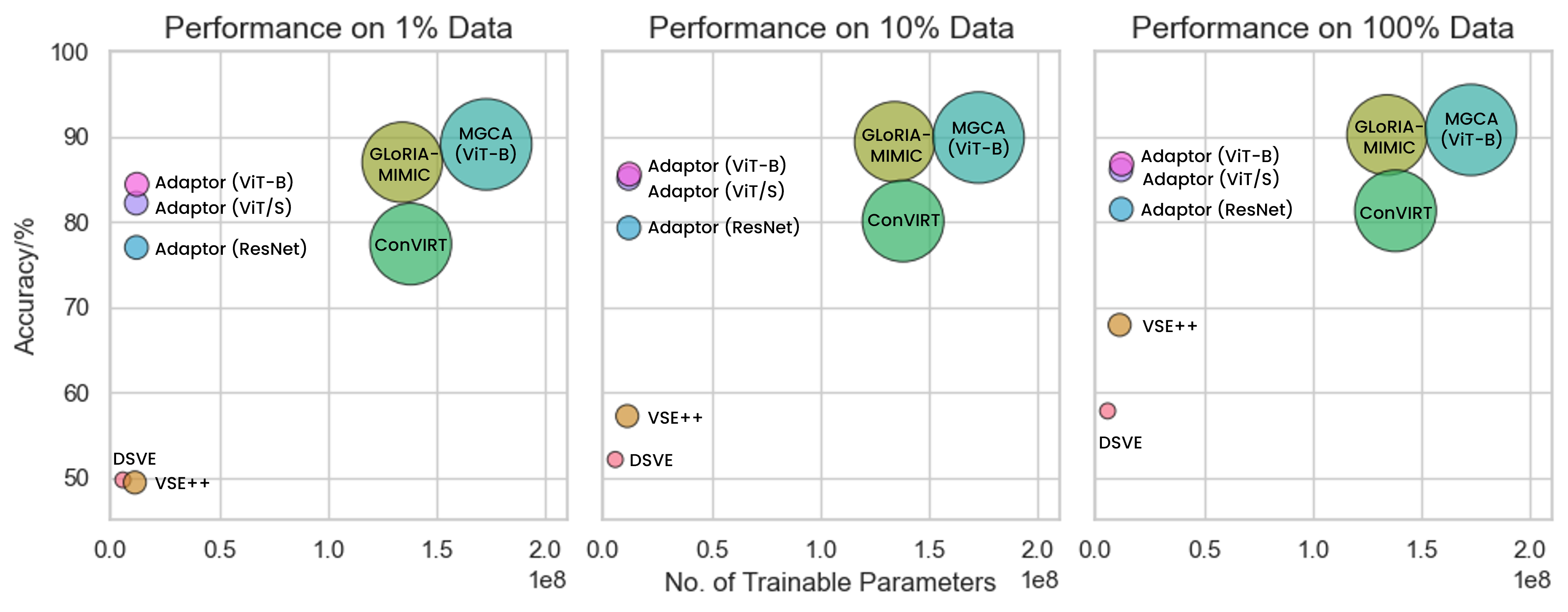}
% \vspace*{-10mm}
  \caption{{{Number of trainable parameters v.s. performance on RSNA classification. The size of the data points also reflect the number of trainable parameters. }}}
  \label{fig:performance-vs-params}
\end{minipage}%
\begin{minipage}{.4\textwidth}
  \centering
  \includegraphics[width=\textwidth]{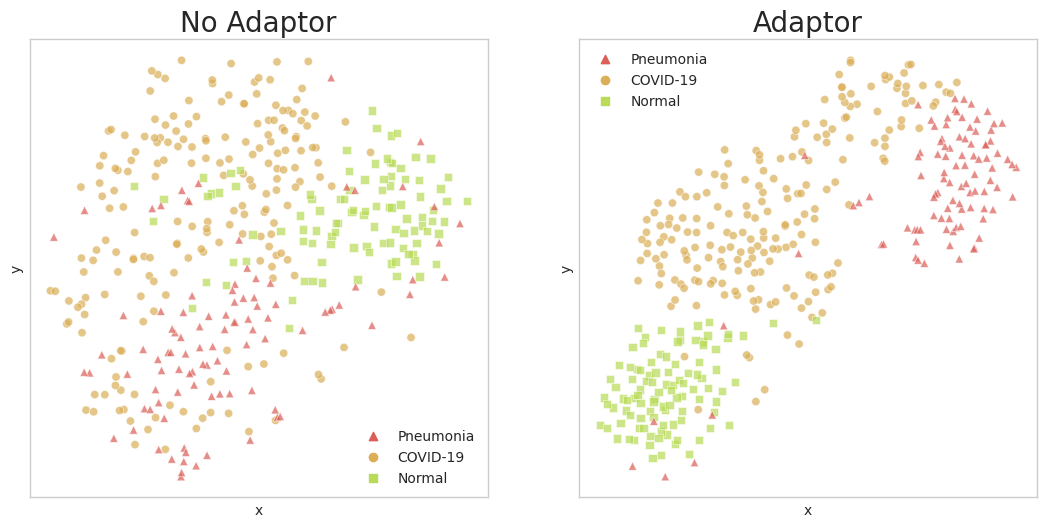}
\vspace*{-5mm}
  \caption{{{ T-SNE Visualisation of vision embeddings from the COVIDx test dataset, before and after Adaptor module. }}}
  \label{fig:tsne}
\end{minipage}
\end{figure*}

\noindent\textbf{Parameter efficiency.} The standout characteristic of our Adaptor framework is its low computational requirement for training. In \Cref{tab:compare-clf-params}, we present parameter counts alongside the performance of various Vision-Language learning methods for classification tasks. Compared with recent medical VL-SSL approaches utilising large multi-modal encoders and trained end-to-end, the Adaptor framework offers competitive results with a reduction of over $90\%$ in trainable parameters. \Cref{fig:performance-vs-params} illustrates the correlation between the classification AUROC score and the number of pre-trained parameters for different methods on the RSNA dataset. While there is a roughly linear correlation between the number of trainable parameters and performance metrics for existing models, the Adaptor framework escapes this linear relationship, achieving performance comparable to large contemporary models with a similar size of trainable parameters as shallow baselines. Utilising the full set of training data, Adaptor with DINOv2-base backbone outperforms ConVIRT \cite{zhang2022contrastive}, with just 8.8\% of its parameters to update during training. \Cref{tab:seg-param-perf} contrasts trainable parameters and performance on the segmentation task. Our Vision-Transformer-based Adaptors, which are set up with a decoder similar in size to previous works, considerably surpass competitors performance trained on full datasets. The ResNet-based Adaptor, although paired with a decoder about twice the size due to its use of ResNet-101 architecture instead of ResNet-50, displays even stronger performance in low-data scenario, achieving superior results with only 1\% of the data compared to its peers trained on the full dataset.

\noindent\textbf{Cross-modal fusion.} 
The Adaptor module is able to learn multi-modal dependencies from the fusion operations, which transfers to enhanced downstream task performance. Using the unseen test split of the COVIDx dataset, we visualise in \Cref{fig:tsne} the t-SNE representation of vision embeddings before and after Adaptor processing. In both cases, there is only a vision backbone involved, as the text encoder has been detached at this stage. However, pre-trained textual and cross-modal knowledge has been stored in the weights of the Adaptor module from pre-training. 
While the pre-trained vision backbone itself struggles to clearly distinguish each class, the enhancements of Adaptor make the three clusters almost linearly separable. It is worth noticing that the Adaptor never received training on classification task signals. This is achieved solely by combining the prior information from the text encoder with vision backbone during pre-training, and directly applying this multi-modal knowledge to downstream tasks, showcasing the effectiveness of the fusion.  

\section{Conclusion}
    We present the Adaptor framework, a parameter-efficient Vision-language Self-Supervised Learning method for enhanced medical vision representation learning. The Adaptor framework freezes pre-trained dual encoders and deploys a backbone-agnostic module with cross-attention for inter-modality fusion. This approach is computationally efficient, preserves the depth of medical knowledge from each individual encoder and synergises them to curate enriched, general-purpose medical features. 
    Our method rivals the performance of recent advanced methodologies while maintaining its training requirements at the level of shallow baselines. Empirical evaluations on two medical imaging tasks across three datasets demonstrate the competency of our framework: it achieves competitive results against existing medical vision-language pre-training approaches, while cutting down training costs and reducing tuneable parameters by over 90\%. 
\newpage
% \section{REFERENCES}
% \label{sec:refs}

% List and number all bibliographical references at the end of the
% paper. The references can be numbered in alphabetic order or in
% order of appearance in the document. When referring to them in
% the text, type the corresponding reference number in square
% brackets as shown at the end of this sentence \cite{li2018attention}. An
% additional final page (the fifth page, in most cases) is
% allowed, but must contain only references to the prior
% literature.

% \bibliographystyle{IEEEbib}
% \bibliography{strings,refs,unsrt}
\ninept
\bibliographystyle{IEEE.bst}
\bibliography{refs.bib}
\end{document}